\pdfoutput=1

\documentclass[11pt]{article}

\usepackage[preprint]{acl}

\usepackage{times}
\usepackage{latexsym}

\usepackage[T1]{fontenc}

\usepackage[utf8]{inputenc}

\usepackage{microtype}

\usepackage{inconsolata}

\usepackage{graphicx}
\usepackage{amsmath}
\usepackage{amssymb}
\usepackage{subcaption}
\usepackage{enumitem}
\usepackage{url}
\usepackage{listings}
\usepackage{xcolor}

\title{A Measure of the System Dependence of Automated Metrics}

\author{Pius von D\"{a}niken 
\and Jan Milan Deriu  
\and Mark Cieliebak \\
Centre for Artificial Intelligence\\
ZHAW School of Engineering \\
\texttt{\{vode,deri,ciel\}@zhaw.ch} \\
}

\begin{document}
\maketitle
\begin{abstract}

Automated metrics for Machine Translation have made significant progress, with the goal of replacing expensive and time-consuming human evaluations. These metrics are typically assessed by their correlation with human judgments, which captures the monotonic relationship between human and metric scores. However, we argue that it is equally important to ensure that metrics treat all systems fairly and consistently. In this paper, we introduce a method to evaluate this aspect.

\end{abstract}

\section{Introduction}
Recent years have seen significant advances in machine translation (MT), marked notably by the introduction of the transformer architecture~\citep{transformer}. Current large-scale commercial systems such as GPT~\citep{gpt3} continue this trend and show promising results~\citep{wmt_general_23, how_good_gpt_mt, lan_bridge_mt}.
A critical supplement to these advancements is thorough and reliable evaluation procedures, which are essential not only for measuring overall progress but also for effectively comparing different systems. While evaluation based on human raters is still considered the gold standard, it is expensive and time-intensive. Therefore, considerable efforts have been made to develop automated metrics for assessing translation quality. Notably, the WMT Metrics series of shared tasks are dedicated to this purpose~\citep[][i.a.]{wmt_metrics_23, wmt_metrics_22, wmt_metrics_21}.
Automated metrics usually assign a scalar~\footnote{Other types of ratings exist, in particular preference ratings~\citep{belz-pref}.} quality rating to a candidate translation based on the source segment and a reference translation. A system-level rating is derived by averaging the segment ratings over a test set. 

To measure a metric's usefulness, we usually measure two aspects: its correlation to human judgments on the segment-level (which checks if there is a monotonic function between metric ratings and human ratings) and whether the system-level ratings can reproduce the same ranking as human ratings~\citep{sign_accuracy, faviscore}. In this paper, we argue that this evaluation of metrics is insufficient, as it ignores a central requirement, namely, that it should treat all systems under evaluation equally. As stated more colloquially, a measuring stick should not change length depending on the measured object. However, this is exactly what we observe in current metrics. 

\begin{figure}[t!]
    \centering
    \includegraphics[width=0.7\linewidth]{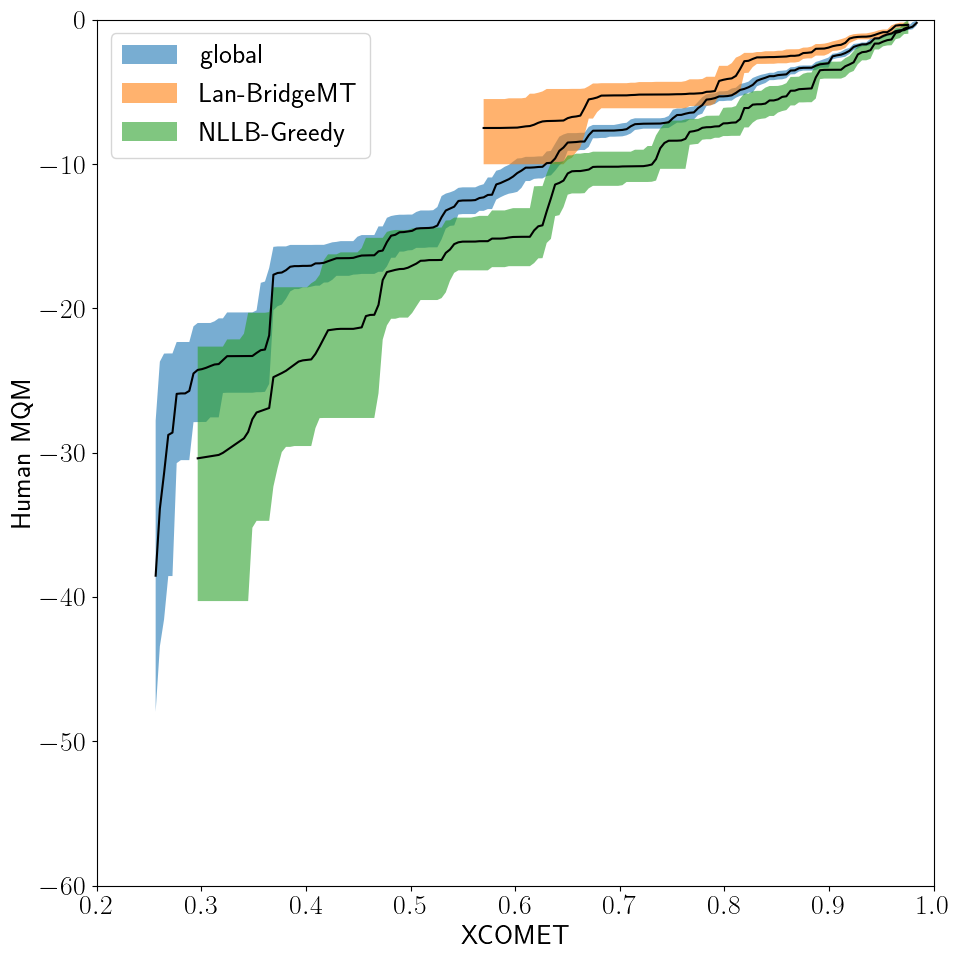} \hfill
    \caption{Average Human Ratings associated with \emph{XCOMET} scores on Chinese to English (\emph{zh-en}) WMT 23 data. We show scores for all system in aggregate (global) and two individual systems.}
    \label{fig:calibration-differences}
\end{figure}

Consider Figure~\ref{fig:calibration-differences}, which shows the expected human rating for each score of the \emph{XCOMET} metric (the best metric in the WMT23 metrics task, with a very high segment-level correlation of 0.65 for the zh-en language pair)~\citep{wmt_metrics_23}. That is, for each possible value that \emph{XCOMET} may assume, we show the expected human rating and the 95\% confidence interval (computed using  Isotonic Regression and bootstrap sampling; see Sections~\ref{sec:theory} and~\ref{sec:experiment} for the details). The global curve (blue) shows the average human score for each metric score if computed over all systems under evaluation in the WMT23 dataset (in standard correlation to human judgment evaluation, we only measure whether this curve is monotonic). In contrast, \emph{Lan-BridgeMT} (best system according to humans) and \emph{NLB-Greedy} (lowest-rated system according to humans) show the average human score for each metric score when computed on one separate system only.  For instance, an \emph{XCOMET} score of 0.7 corresponds to an average Human-MQM score of $-5.2$ for \emph{Lan-BridgeMT}, and Human-MQM score of $-10.2$ for \emph{NLB-Greedy}. 

This leads to the following consequence: For \emph{Lan-BridgeMT}, higher human scores are associated with lower metric scores than the global curve, which leads to an underestimation of \emph{Lan-BridgeMT}'s performance, according to \emph{XCOMET}. The opposite effect is visible for \emph{NLB-Greedy}, which is overestimated and, in fact, gains 3 ranks (from 15th to 12th place) when comparing the metric and human ranking (see also Table~\ref{tab:wmt23-zhen-xcomet-scores}  in Section~\ref{sec:experiment}). Thus, a metric that exhibits a high global segment-level correlation to human judgments can lead to wrong system-level rankings. This observation leads us to the central claim of this paper: \textbf{The cause of the discrepancy between the correlation on the segment level and the final system ranking is due to the metric's dependency of the system under evaluation.} 


The main position of this paper is that when evaluating a novel metric, one ought to measure the dependency on the system under evaluation as well, alongside the correlation to human judgment. In the following, we will formalize this dependency of the relation between human and metric ratings on the system under evaluation and derive a measure for quantifying this effect.

\section{Averaging Metric Scores}\label{sec:theory}

Assume we are given a set of $K$ machine translation systems $\pi_k$ to evaluate. A translation system maps an input sentence $i \in \mathcal{I}$ in a fixed source language to an output sentence $o \in \mathcal{O}$ in a fixed target language: $\pi_k: \mathcal{I} \rightarrow \mathcal{O}$. The usual human evaluation scenario involves curating a test set of $N$ inputs $\mathcal{T} = \left\{ i^{(j)} | 1 \le j \le N\right\} \subset \mathcal{I}$ for which we collect the output of each system $\pi_k$ for each input $i \in \mathcal{T}$, and then ask human annotators to produce ratings. This results in a set of ratings $\left\{ (h_{1}^{(j)}, \dots, h_{k}^{(j)}, \dots, h_{K}^{(j)}) | 1 \le j \le N \right\}$, where $h_{k}^{(j)} \in \mathbb{R}$ is a scalar rating provided by human annotators measuring the quality of the translation provided by $\pi_k$ for input $i^{(j)}$. We will assume that higher human ratings indicate higher translation quality. In this setting, it is natural to measure the overall quality of system $\pi_k$ by the average human rating it achieves $\hat{\mu}_{k}^{H} = \frac{1}{N}\sum_{j=1}^{N}h_{k}^{(j)}$. This is an estimator of the expected human rating $\mu_{k}^{H} = \mathbb{E}[h_{k}]$ achieved by $\pi_k$ for any input in $\mathcal{I}$, assuming that $\mathcal{T}$ is appropriately chosen.

In many cases, we want to replace human raters with an automated scalar metric $M: \mathcal{I} \times \mathcal{O} \rightarrow \mathbb{R}$, which maps an input and translation to a scalar value. For our test set $\mathcal{T}$, we can collect all metric ratings $\left\{ (m_{1}^{(j)}, \dots, m_{k}^{(j)}, \dots, m_{K}^{(j)}) | 1 \le j \le N \right\}$, where $m_{k}^{(j)} = M(i^{(j)}, \pi_k(i^{(j)}))$, the metric rating for input $i^{(j)}$ and translation by $\pi_k$. In this case, it is common to use the sample average $\hat{\mu}_{k}^{M} = \frac{1}{N}\sum_{j=1}^{N}m_{k}^{(j)}$ to measure the quality of system $\pi_k$, which is an estimator of the expected metric rating $\mu_{k}^{M} = \mathbb{E}[m_k]$ achieved by $\pi_k$.

The goal of automated evaluation is to use $\hat{\mu}_{k}^{M}$ as a proxy measure for $\mu_{k}^{H}$, in particular, to rank the systems $\pi_1, \dots, \pi_K$ according to their performance. In the following, we will study the relationship between $\mu_{k}^{H}$ and $\mu_{k}^{M}$, which is expressed by an unknown function $f_G$ that maps from the metric scale to the human scale. There are two requirements to this function: first, that it is monotonic (i.e., that it respects the order of the metric scale), and second, that it does not depend on the system under evaluation $\pi_{k}$ (i.e., that it is the same for all systems) The goal is to find the relation between $\mu_{k}^{H}$ and  $\mu_{k}^{M}$. The idea is to express $\mathbb{E}[h_k]$ in terms of an expectation over metric ratings as follows (for full derivation, see Appendix~\ref{sec:full_derivation}): 

\begin{align}
\small
\begin{split}
    \label{eq:cond_expectation}
    \mathbb{E}[h_k] = \mathbb{E}_{p_k(m)}[\mathbb{E}_{p_k(h)}[h | m]]
\end{split}
\end{align}

The crucial element of Equation~\ref{eq:cond_expectation} is the conditional expectation $\mathbb{E}_{p_k(h)}[h | m]$. Here we consider the expectation according to $p_k(h)$, the distribution of human ratings for system $\pi_k$. Equation~\ref{eq:cond_expectation} describes the relationship between $\mu_{k}^{H}$ and $\mu_{k}^{M}$ by expressing the expected human rating in terms of an expectation over metric ratings. We interpret this element as a function $f_k$, which takes a metric rating as input and returns the expected human rating.
Equation~\ref{eq:cond_expectation} yields a function $f_k$ for each system separately, which is not necessarily the same across systems. At this point, we can restate the introductory discussion using our formalism. When averaging metrics $\hat{\mu}_{k}^{M}$ to rank systems, we implicitly assume that there is a global function $f_G$ that is equal to all the system-specific functions $f_k$, i.e., $f_G = f_1 = \dots = f_K$, and thus, only measure if $f_G$ is monotonic (through correlation to human judgments). However, as shown in Figure~\ref{fig:calibration-differences}, this assumption does not hold in practice (where blue is $f_G$, and we have an $f_k$ for the two other systems respectively). 
To show that this is insufficient, we consider the effects of violating the assumption. Let us assume $f_1 \neq f_2$, but both are monotonic. Consider the extreme example that $\mu_1^{M} = \mu_2^{M}$, i.e., systems $\pi_1$ and $\pi_2$ are of the same quality under the metric. However, consider the case $f_1(m) = f_2(m) + C$, $C > 0$. Then $\frac{1}{N}\sum_{j}f_1(m_{1}^{(j)}) = C + \frac{1}{N}\sum_{j}f_2(m_{1}^{(j)}) > \frac{1}{N}\sum_{j}f_2(m_{2}^{(j)})$, thus, yielding that $\pi_1$ is better than $\pi_2$ in human space. This shows the necessity of measuring the monotonicity of a global function $f_G$ and the dependence of the metric on the systems under evaluation.

We first introduce the Expected Deviation (ED), which measures the difference between $f_G$ and $f_k$ for all $k \in \{1 \dots K\}$, which tells us how much a system is over-or-underestimated according to the metric. That is the difference 
\begin{align}
\small
\begin{split}
\label{eq:score}
   \textit{ED}(k) &= \frac{1}{N}\sum_{j=1}^{N}f_{G}(m_{k}^{(j)}) -  \frac{1}{N}\sum_{j=1}^{N}f_{k}(m_{k}^{(j)})
\end{split}
\end{align}
This is equivalent to $\mu^G_k - \mu^H_k $, where $\mu^G_k  =  \frac{1}{N}\sum_{j=1}^{N}f_{G}(m_{k}^{(j)})$, thus, we measure the difference between the average rating according to the global function and the average rating of the system-specific function, which corresponds to the human rating-average. Note that a mis-ranking occurs if one system is severely overrated while another is severely underrated. Thus we define the system dependence score  $\textit{SysDep}(\mathcal{M})$ as the worst case of this effect:
\begin{align}
\small
\begin{split}
\label{eq:score}
   \textit{SysDep}(\mathcal{M}) &= max_{\pi_k} ED(k) - min_{\pi_k} ED(k)
\end{split}
\end{align}



\section{Experiments}\label{sec:experiment}

\paragraph{Estimating the Conditional Expectation.} Even though the functions $f_k$ and $f_G$ are unknown in general, we can estimate them from data. We will use \emph{Isotonic Regression (IR)}~\citep{isotonic-regression} for this purpose, which estimates a monotonic function $\hat{f}_k$ minimizing $\sum_{j}(\hat{f}_k(m_{k}^{(j)}) - h_{k}^{(j)})^{2}$. To estimate $f_G$, we utilize the same approach, pooling all paired data from all systems. 
To compute the \emph{SysDep} of a metric, we compute the ED for each MT system under that metric. For this, we compute the average human rating $\hat{\mu}_{k}^{H} = \frac{1}{N_H}\sum_{j=1}^{N_H}h_{k}^{(j)}$, the average metric rating $\hat{\mu}_{k}^{M} = \frac{1}{N_M}\sum_{j=1}^{N_M}m_{k}^{(j)}$, as well as average remapped rating $\hat{\mu}_{k}^{G} = \frac{1}{N_M}\sum_{j=1}^{N_M}\hat{f}_{G}(m_{k}^{(j)})$ for each MT system. We provide our code in Appendix~\ref{sec:code}.

\begin{table*}[tb]
    \centering
    \small
    \begin{tabular}{r | c c c c c c c}
 & \multicolumn{2}{c}{Human} & \multicolumn{2}{c}{Metric} & \multicolumn{2}{c}{Remapped} & Exp. Deviation\\
 & $\hat{\mu}_{k}^{H}$ & R & $\hat{\mu}_{k}^{M}$ & R & $\hat{\mu}_{k}^{G}$ & R & ED \\ \hline
Lan-BridgeMT & -2.100 & 1 & 0.889 & 2 & -2.920 & 2 & \textit{-0.820} \\
GPT4-5shot & -2.305 & 2 & 0.893 & 1 & -2.800 & 1 & -0.494 \\
Yishu & -3.231 & 3 & 0.880 & 4 & -3.179 & 4 & 0.052 \\
ONLINE-B & -3.385 & 4 & 0.879 & 5 & -3.188 & 5 & 0.197 \\
HW-TSC & -3.398 & 5 & 0.883 & 3 & -3.080 & 3 & 0.318 \\
ONLINE-A & -3.785 & 6 & 0.856 & 8 & -3.812 & 8 & -0.027 \\
ONLINE-Y & -3.792 & 7 & 0.868 & 6 & -3.479 & 6 & 0.313 \\
ONLINE-G & -3.857 & 8 & 0.864 & 7 & -3.607 & 7 & 0.250 \\
ONLINE-W & -4.062 & 9 & 0.848 & 9 & -4.165 & 10 & -0.103 \\
ZengHuiMT & -4.232 & 10 & 0.846 & 10 & -4.140 & 9 & 0.092 \\
IOL-Research & -4.586 & 11 & 0.843 & 11 & -4.251 & 11 & 0.335 \\
ONLINE-M & -5.433 & 12 & 0.820 & 15 & -4.907 & 15 & 0.526 \\
ANVITA & -6.078 & 13 & 0.830 & 13 & -4.602 & 13 & 1.475 \\
NLLB-MBR-BLEU & -6.360 & 14 & 0.825 & 14 & -4.726 & 14 & 1.634 \\
NLLB-Greedy & -6.574 & 15 & 0.831 & 12 & -4.578 & 12 & \textbf{1.996} \\
\end{tabular}

    \caption{System rankings and average rating of WMT 23 \emph{zh-en} systems according to \emph{XCOMET}. The lowest score is in italics, and the highest is in bold.}
    \label{tab:wmt23-zhen-xcomet-scores}
\end{table*}

\paragraph{Data.} We rely on data from the WMT 23 Metrics shared task~\citep{wmt_metrics_23}. The data includes translations for 3 language pairs: English to German (\emph{en-de}), Hebrew to English (\emph{he-en}), and Chinese to English (\emph{zh-en}). The translations were produces by 12-15 systems (depending on the language pair) which participated in the general MT task~\citep{wmt_general_23}. Human ratings are available in the form of MQM annotations~\citep{mqm}, which are based on error-span annotations by experts that are subsequently transformed into a numeric value by assigning scores to errors based on their severity. Here, we will present results for the \emph{XCOMET}~\citep{xcomet} metric (best metric according to correlation to human judgments) and the \emph{zh-en} language pair, where we have access to $N_M = 1976$ segments per system rated by the metric and $N_H = 1177$ of these segments rated with human MQM ratings. Results for the other language pairs and an additional metric are shown in Appendix~\ref{app:additional-results}. To estimate the conditional expectation functions $f_k$, we use the $1177$ paired ratings for each system $\pi_k$. We employ $B = 200$ bootstrap samples of the paired data to fit $B$ IR models. Our estimate, $\hat{f}_k$, represents the average of these $B$ IR models. In Figure~\ref{fig:calibration-differences}, we also present the range between the 2.5\% and 97.5\% percentiles. 

\paragraph{Results.} We show the results in Table~\ref{tab:wmt23-zhen-xcomet-scores}. We can see that the ED ranges from -0.82 to 1.996, thus yielding a \emph{SysDep} score of 2.816. We see that both \emph{Lan-BridgeMT} and \emph{GPT4-5shot} are underrated by the metric (negative \textit{ED}), but \emph{Lan-BridgeMT} more so, enough to invert their order. At the bottom of the ranking, we see a relatively large absolute \textit{ED}. Ranking errors reflect an interplay between the systems' rating gap and the \textit{ED}s. For example, \emph{Online-A} loses 2 ranks according to the metric even though it has the lowest absolute \textit{ED}. We also note that even though $\hat{f}_G$ is monotonic, the ranking between the metric and the remapped scores does not match completely. This can be attributed to the uncertainty introduced by bootstrapping and extrapolating to the unpaired metric ratings. It can be seen for \emph{ONLINE-W} and \emph{ZengHuiMT}, which have similar metric ratings. 

Overall, our results show that although there is a highly monotonic function between the \emph{XCOMET} scale and the human scale, \emph{XCOMET} exhibits a high dependency on the system under evaluation, thus yielding an inconsistent ranking between humans and \emph{XCOMET}.   

\section{Related Work}\label{sec:related}

The derivation in Section~\ref{sec:theory} closely follows \citet{calibrate_extrapolate}, who provide the same argument in the context of binary prevalence estimation. In our case, the conditional expectation $\mathbb{E}[h |m]$ plays the same role as the calibration curve in their framework. Under that lens, the Expected Deviation is analogous to the Expected Calibration Error~\citep{expected-calibration-error}. Following the same analogy, evaluating a new MT system is similar to applying a classifier to a new domain.

Previous studies by \citet{ours_pref} and \citet{ours_binary} have highlighted that metric performance depends on the system under test. They employed a Bayesian framework to determine the proportions of binary or preference human ratings from metric scores; critically relying on confusion matrices estimated for each MT system. In this discrete rating context, these confusion matrices represent the same concept as $\mathbb{E}[h|m]$. In follow-up work, \citet{faviscore} find that some metrics disproportionately favor certain MT systems over others compared to human preference ratings. Our finding provides a plausible explanation.

\citet{debiasing} shows how to combine human ratings and metric ratings to derive an unbiased estimate of the true expected human rating $\mu^{H}$ while reducing the number of annotations needed. The proposed control variates estimator is based only on human and metric scores for a given MT system, even when estimating their correlation, thus avoiding the problem we describe.

\citet{statistical_advantage} consider disagreements in the ordering of systems when using $\mu_{k}^{M}$ instead of $\mu_{k}^{H}$. In particular they study the sign error, cases where $sign(\mu_{1}^{M} - \mu_{2}^{M}) \ne sign(\mu_{1}^{H} - \mu_{2}^{H})$. They apply a bias variance decomposition to this error and find that while the human estimator is unbiased, it exhibits high variance while the opposite is the case for metrics. Our \textit{SysDep} score presents a way to quantify this bias. 


\section{Conclusion}

In this paper, we emphasize the importance of ensuring that automated metrics treat all MT systems consistently, a factor overlooked in current evaluations. By mapping metric scores to the human rating scale, we estimate how much a metric misjudges individual system performance. We compute the range of these deviations to assess how consistently a metric treats different systems. In Appendix~\ref{sec:sysdep}, we re-evaluate WMT23 metrics from this perspective. Additionally, in Appendix~\ref{sec:intra-system}, we confirm that these results stem from systematic differences in how metrics treat systems by measuring deviations within splits of a single system's ratings.

\section*{Limitations}

This paper is intended to explore an overlooked aspect of the evaluation of automated metrics. The \textit{SysDep} measure we developed will hopefully provide a starting point for the development of more refined evaluation of the way metrics treat different systems differently.

Our experiments are based solely on data from the WMT23 Metrics shared task. To further solidify our findings a larger scale study with more domains and larger sample sizes are needed.

While we provide a way to measure the system dependence of a metric, we do not provide any suggestions on how to develop metrics that minimize the \textit{SysDep}.

\section*{Acknowledgments}
This work was supported by the Swiss National Science Foundation (SNF) within the project "Unified Model for Evaluation of Text Generation Systems (UniVal)" [200020\_219819].

\bibliography{custom}

\begin{thebibliography}{26}
\providecommand{\natexlab}[1]{#1}

\bibitem[{Barlow and Brunk(1972)}]{isotonic-regression}
R.~E. Barlow and H.~D. Brunk. 1972.
\newblock \href {https://doi.org/10.1080/01621459.1972.10481216} {The isotonic regression problem and its dual}.
\newblock \emph{Journal of the American Statistical Association}, 67(337):140--147.

\bibitem[{Belz and Kow(2010)}]{belz-pref}
Anja Belz and Eric Kow. 2010.
\newblock \href {https://aclanthology.org/W10-4201} {Comparing rating scales and preference judgements in language evaluation}.
\newblock In \emph{Proceedings of the 6th International Natural Language Generation Conference}. Association for Computational Linguistics.

\bibitem[{Brown et~al.(2020)Brown, Mann, Ryder, Subbiah, Kaplan, Dhariwal, Neelakantan, Shyam, Sastry, Askell, Agarwal, Herbert-Voss, Krueger, Henighan, Child, Ramesh, Ziegler, Wu, Winter, Hesse, Chen, Sigler, Litwin, Gray, Chess, Clark, Berner, McCandlish, Radford, Sutskever, and Amodei}]{gpt3}
Tom~B. Brown, Benjamin Mann, Nick Ryder, Melanie Subbiah, Jared Kaplan, Prafulla Dhariwal, Arvind Neelakantan, Pranav Shyam, Girish Sastry, Amanda Askell, Sandhini Agarwal, Ariel Herbert-Voss, Gretchen Krueger, Tom Henighan, Rewon Child, Aditya Ramesh, Daniel~M. Ziegler, Jeffrey Wu, Clemens Winter, Christopher Hesse, Mark Chen, Eric Sigler, Mateusz Litwin, Scott Gray, Benjamin Chess, Jack Clark, Christopher Berner, Sam McCandlish, Alec Radford, Ilya Sutskever, and Dario Amodei. 2020.
\newblock Language models are few-shot learners.
\newblock In \emph{Proceedings of the 34th International Conference on Neural Information Processing Systems}, NIPS '20, Red Hook, NY, USA. Curran Associates Inc.

\bibitem[{Chaganty et~al.(2018)Chaganty, Mussmann, and Liang}]{debiasing}
Arun Chaganty, Stephen Mussmann, and Percy Liang. 2018.
\newblock \href {https://doi.org/10.18653/v1/P18-1060} {The price of debiasing automatic metrics in natural language evalaution}.
\newblock In \emph{Proceedings of the 56th Annual Meeting of the Association for Computational Linguistics (Volume 1: Long Papers)}, pages 643--653, Melbourne, Australia. Association for Computational Linguistics.

\bibitem[{Deriu et~al.(2023)Deriu, von D{\"a}niken, Tuggener, and Cieliebak}]{ours_pref}
Jan Deriu, Pius von D{\"a}niken, Don Tuggener, and Mark Cieliebak. 2023.
\newblock \href {https://doi.org/10.18653/v1/2023.findings-acl.404} {Correction of errors in preference ratings from automated metrics for text generation}.
\newblock In \emph{Findings of the Association for Computational Linguistics: ACL 2023}, pages 6456--6474, Toronto, Canada. Association for Computational Linguistics.

\bibitem[{Freitag et~al.(2023)Freitag, Mathur, Lo, Avramidis, Rei, Thompson, Kocmi, Blain, Deutsch, Stewart, Zerva, Castilho, Lavie, and Foster}]{wmt_metrics_23}
Markus Freitag, Nitika Mathur, Chi-kiu Lo, Eleftherios Avramidis, Ricardo Rei, Brian Thompson, Tom Kocmi, Frederic Blain, Daniel Deutsch, Craig Stewart, Chrysoula Zerva, Sheila Castilho, Alon Lavie, and George Foster. 2023.
\newblock \href {https://doi.org/10.18653/v1/2023.wmt-1.51} {Results of {WMT}23 metrics shared task: Metrics might be guilty but references are not innocent}.
\newblock In \emph{Proceedings of the Eighth Conference on Machine Translation}, pages 578--628, Singapore. Association for Computational Linguistics.

\bibitem[{Freitag et~al.(2022)Freitag, Rei, Mathur, Lo, Stewart, Avramidis, Kocmi, Foster, Lavie, and Martins}]{wmt_metrics_22}
Markus Freitag, Ricardo Rei, Nitika Mathur, Chi-kiu Lo, Craig Stewart, Eleftherios Avramidis, Tom Kocmi, George Foster, Alon Lavie, and Andr{\'e} F.~T. Martins. 2022.
\newblock \href {https://aclanthology.org/2022.wmt-1.2} {Results of {WMT}22 metrics shared task: Stop using {BLEU} {--} neural metrics are better and more robust}.
\newblock In \emph{Proceedings of the Seventh Conference on Machine Translation (WMT)}, pages 46--68, Abu Dhabi, United Arab Emirates (Hybrid). Association for Computational Linguistics.

\bibitem[{Freitag et~al.(2021)Freitag, Rei, Mathur, Lo, Stewart, Foster, Lavie, and Bojar}]{wmt_metrics_21}
Markus Freitag, Ricardo Rei, Nitika Mathur, Chi-kiu Lo, Craig Stewart, George Foster, Alon Lavie, and Ond{\v{r}}ej Bojar. 2021.
\newblock \href {https://aclanthology.org/2021.wmt-1.73} {Results of the {WMT}21 metrics shared task: Evaluating metrics with expert-based human evaluations on {TED} and news domain}.
\newblock In \emph{Proceedings of the Sixth Conference on Machine Translation}, pages 733--774, Online. Association for Computational Linguistics.

\bibitem[{Guerreiro et~al.(2023)Guerreiro, Rei, van Stigt, Coheur, Colombo, and Martins}]{xcomet}
Nuno~M. Guerreiro, Ricardo Rei, Daan van Stigt, Luisa Coheur, Pierre Colombo, and André F.~T. Martins. 2023.
\newblock \href {https://arxiv.org/abs/2310.10482} {xcomet: Transparent machine translation evaluation through fine-grained error detection}.
\newblock \emph{Preprint}, arXiv:2310.10482.

\bibitem[{Harris et~al.(2020)Harris, Millman, van~der Walt, Gommers, Virtanen, Cournapeau, Wieser, Taylor, Berg, Smith, Kern, Picus, Hoyer, van Kerkwijk, Brett, Haldane, del R{\'{i}}o, Wiebe, Peterson, G{\'{e}}rard-Marchant, Sheppard, Reddy, Weckesser, Abbasi, Gohlke, and Oliphant}]{numpy}
Charles~R. Harris, K.~Jarrod Millman, St{\'{e}}fan~J. van~der Walt, Ralf Gommers, Pauli Virtanen, David Cournapeau, Eric Wieser, Julian Taylor, Sebastian Berg, Nathaniel~J. Smith, Robert Kern, Matti Picus, Stephan Hoyer, Marten~H. van Kerkwijk, Matthew Brett, Allan Haldane, Jaime~Fern{\'{a}}ndez del R{\'{i}}o, Mark Wiebe, Pearu Peterson, Pierre G{\'{e}}rard-Marchant, Kevin Sheppard, Tyler Reddy, Warren Weckesser, Hameer Abbasi, Christoph Gohlke, and Travis~E. Oliphant. 2020.
\newblock \href {https://doi.org/10.1038/s41586-020-2649-2} {Array programming with {NumPy}}.
\newblock \emph{Nature}, 585(7825):357--362.

\bibitem[{Hendy et~al.(2023)Hendy, Abdelrehim, Sharaf, Raunak, Gabr, Matsushita, Kim, Afify, and Awadalla}]{how_good_gpt_mt}
Amr Hendy, Mohamed Abdelrehim, Amr Sharaf, Vikas Raunak, Mohamed Gabr, Hitokazu Matsushita, Young~Jin Kim, Mohamed Afify, and Hany~Hassan Awadalla. 2023.
\newblock \href {https://arxiv.org/abs/2302.09210} {How good are gpt models at machine translation? a comprehensive evaluation}.
\newblock \emph{Preprint}, arXiv:2302.09210.

\bibitem[{Juraska et~al.(2023)Juraska, Finkelstein, Deutsch, Siddhant, Mirzazadeh, and Freitag}]{metricx}
Juraj Juraska, Mara Finkelstein, Daniel Deutsch, Aditya Siddhant, Mehdi Mirzazadeh, and Markus Freitag. 2023.
\newblock \href {https://doi.org/10.18653/v1/2023.wmt-1.63} {{M}etric{X}-23: The {G}oogle submission to the {WMT} 2023 metrics shared task}.
\newblock In \emph{Proceedings of the Eighth Conference on Machine Translation}, pages 756--767, Singapore. Association for Computational Linguistics.

\bibitem[{Kocmi et~al.(2023)Kocmi, Avramidis, Bawden, Bojar, Dvorkovich, Federmann, Fishel, Freitag, Gowda, Grundkiewicz, Haddow, Koehn, Marie, Monz, Morishita, Murray, Nagata, Nakazawa, Popel, Popovi{\'c}, and Shmatova}]{wmt_general_23}
Tom Kocmi, Eleftherios Avramidis, Rachel Bawden, Ond{\v{r}}ej Bojar, Anton Dvorkovich, Christian Federmann, Mark Fishel, Markus Freitag, Thamme Gowda, Roman Grundkiewicz, Barry Haddow, Philipp Koehn, Benjamin Marie, Christof Monz, Makoto Morishita, Kenton Murray, Makoto Nagata, Toshiaki Nakazawa, Martin Popel, Maja Popovi{\'c}, and Mariya Shmatova. 2023.
\newblock \href {https://doi.org/10.18653/v1/2023.wmt-1.1} {Findings of the 2023 conference on machine translation ({WMT}23): {LLM}s are here but not quite there yet}.
\newblock In \emph{Proceedings of the Eighth Conference on Machine Translation}, pages 1--42, Singapore. Association for Computational Linguistics.

\bibitem[{Kocmi and Federmann(2023)}]{gemba-mqm}
Tom Kocmi and Christian Federmann. 2023.
\newblock \href {https://doi.org/10.18653/v1/2023.wmt-1.64} {{GEMBA}-{MQM}: Detecting translation quality error spans with {GPT}-4}.
\newblock In \emph{Proceedings of the Eighth Conference on Machine Translation}, pages 768--775, Singapore. Association for Computational Linguistics.

\bibitem[{Kocmi et~al.(2021)Kocmi, Federmann, Grundkiewicz, Junczys-Dowmunt, Matsushita, and Menezes}]{sign_accuracy}
Tom Kocmi, Christian Federmann, Roman Grundkiewicz, Marcin Junczys-Dowmunt, Hitokazu Matsushita, and Arul Menezes. 2021.
\newblock \href {https://aclanthology.org/2021.wmt-1.57} {To ship or not to ship: An extensive evaluation of automatic metrics for machine translation}.
\newblock In \emph{Proceedings of the Sixth Conference on Machine Translation}, pages 478--494, Online. Association for Computational Linguistics.

\bibitem[{Lommel et~al.(2014)Lommel, Uszkoreit, and Burchardt}]{mqm}
Arle. Language Technology~Lab) Lommel, Hans. Language Technology~Lab) Uszkoreit, and Aljoscha. Language Technology~Lab) Burchardt. 2014.
\newblock \href {https://doi.org/10.5565/rev/tradumatica.77} {Multidimensional quality metrics (mqm) : a framework for declaring and describing translation quality metrics}.
\newblock \emph{Tradumàtica}, (12):455--463.

\bibitem[{Pedregosa et~al.(2011)Pedregosa, Varoquaux, Gramfort, Michel, Thirion, Grisel, Blondel, Prettenhofer, Weiss, Dubourg, Vanderplas, Passos, Cournapeau, Brucher, Perrot, and Duchesnay}]{scikit-learn}
F.~Pedregosa, G.~Varoquaux, A.~Gramfort, V.~Michel, B.~Thirion, O.~Grisel, M.~Blondel, P.~Prettenhofer, R.~Weiss, V.~Dubourg, J.~Vanderplas, A.~Passos, D.~Cournapeau, M.~Brucher, M.~Perrot, and E.~Duchesnay. 2011.
\newblock Scikit-learn: Machine learning in {P}ython.
\newblock \emph{Journal of Machine Learning Research}, 12:2825--2830.

\bibitem[{Posocco and Bonnefoy(2021)}]{expected-calibration-error}
Nicolas Posocco and Antoine Bonnefoy. 2021.
\newblock Estimating expected calibration errors.
\newblock In \emph{Artificial Neural Networks and Machine Learning -- ICANN 2021}, pages 139--150, Cham. Springer International Publishing.

\bibitem[{Thompson and Post(2020{\natexlab{a}})}]{prismA}
Brian Thompson and Matt Post. 2020{\natexlab{a}}.
\newblock \href {https://doi.org/10.18653/v1/2020.emnlp-main.8} {Automatic machine translation evaluation in many languages via zero-shot paraphrasing}.
\newblock In \emph{Proceedings of the 2020 Conference on Empirical Methods in Natural Language Processing (EMNLP)}, pages 90--121, Online. Association for Computational Linguistics.

\bibitem[{Thompson and Post(2020{\natexlab{b}})}]{prismB}
Brian Thompson and Matt Post. 2020{\natexlab{b}}.
\newblock \href {https://aclanthology.org/2020.wmt-1.67} {Paraphrase generation as zero-shot multilingual translation: Disentangling semantic similarity from lexical and syntactic diversity}.
\newblock In \emph{Proceedings of the Fifth Conference on Machine Translation}, pages 561--570, Online. Association for Computational Linguistics.

\bibitem[{Vaswani et~al.(2017)Vaswani, Shazeer, Parmar, Uszkoreit, Jones, Gomez, Kaiser, and Polosukhin}]{transformer}
Ashish Vaswani, Noam Shazeer, Niki Parmar, Jakob Uszkoreit, Llion Jones, Aidan~N. Gomez, \L{}ukasz Kaiser, and Illia Polosukhin. 2017.
\newblock Attention is all you need.
\newblock In \emph{Proceedings of the 31st International Conference on Neural Information Processing Systems}, NIPS'17, page 6000–6010, Red Hook, NY, USA. Curran Associates Inc.

\bibitem[{von D{\"a}niken et~al.(2022)von D{\"a}niken, Deriu, Tuggener, and Cieliebak}]{ours_binary}
Pius von D{\"a}niken, Jan Deriu, Don Tuggener, and Mark Cieliebak. 2022.
\newblock \href {https://doi.org/10.18653/v1/2022.findings-emnlp.108} {On the effectiveness of automated metrics for text generation systems}.
\newblock In \emph{Findings of the Association for Computational Linguistics: EMNLP 2022}, pages 1503--1522, Abu Dhabi, United Arab Emirates. Association for Computational Linguistics.

\bibitem[{von D{\"a}niken et~al.(2024)von D{\"a}niken, Deriu, Tuggener, and Cieliebak}]{faviscore}
Pius von D{\"a}niken, Jan Deriu, Don Tuggener, and Mark Cieliebak. 2024.
\newblock \href {https://doi.org/10.18653/v1/2024.acl-long.243} {Favi-score: A measure for favoritism in automated preference ratings for generative {AI} evaluation}.
\newblock In \emph{Proceedings of the 62nd Annual Meeting of the Association for Computational Linguistics (Volume 1: Long Papers)}, pages 4437--4454, Bangkok, Thailand. Association for Computational Linguistics.

\bibitem[{Wei and Jia(2021)}]{statistical_advantage}
Johnny Wei and Robin Jia. 2021.
\newblock \href {https://doi.org/10.18653/v1/2021.acl-long.533} {The statistical advantage of automatic {NLG} metrics at the system level}.
\newblock In \emph{Proceedings of the 59th Annual Meeting of the Association for Computational Linguistics and the 11th International Joint Conference on Natural Language Processing (Volume 1: Long Papers)}, pages 6840--6854, Online. Association for Computational Linguistics.

\bibitem[{Wu and Resnick(2024)}]{calibrate_extrapolate}
Siqi Wu and Paul Resnick. 2024.
\newblock \href {https://doi.org/10.1609/icwsm.v18i1.31414} {Calibrate-extrapolate: Rethinking prevalence estimation with black box classifiers}.
\newblock \emph{Proceedings of the International AAAI Conference on Web and Social Media}, 18(1):1634--1647.

\bibitem[{Wu and Hu(2023)}]{lan_bridge_mt}
Yangjian Wu and Gang Hu. 2023.
\newblock \href {https://doi.org/10.18653/v1/2023.wmt-1.15} {Exploring prompt engineering with {GPT} language models for document-level machine translation: Insights and findings}.
\newblock In \emph{Proceedings of the Eighth Conference on Machine Translation}, pages 166--169, Singapore. Association for Computational Linguistics.

\end{thebibliography}

\appendix

\section{Full Derivation}
\label{sec:full_derivation}

In Equation~\ref{eq:cond_expectation_der}, we give the full derivation of Equation~\ref{eq:cond_expectation} in Section~\ref{sec:theory}. In the following $p_{k}(h)$ is the density of human ratings for system $\pi_k$, $p_{k}(m)$ is its density of metric ratings, and $p_k(h, m)$ the joint density.

\begin{align}
\small
\begin{split}
\label{eq:cond_expectation_der}
    \mathbb{E}[h_k] &= \int_{-\infty}^{\infty}h p_k(h)\mathrm{d}h \\
    &= \int_{-\infty}^{\infty}h \left[ \int_{-\infty}^{\infty}p_k(h, m)\mathrm{d}m \right]\mathrm{d}h \\
    &= \int_{-\infty}^{\infty}h \left[ \int_{-\infty}^{\infty}p_k(h | m)p_k(m)\mathrm{d}m \right]\mathrm{d}h \\
    &= \int_{-\infty}^{\infty}\int_{-\infty}^{\infty}hp_k(h|m)p_k(m)\mathrm{d}m\mathrm{d}h \\
    &= \int_{-\infty}^{\infty} \left[ \int_{-\infty}^{\infty}hp_k(h|m)\mathrm{d}h \right]p_k(m)\mathrm{d}m \\
    &= \int_{-\infty}^{\infty} \mathbb{E}[h_k | m]p_k(m)\mathrm{d}m \\
    &= \mathbb{E}_{p_k(m)}[\mathbb{E}_{p_k(h)}[h | m]]
\end{split}
\end{align}



\section{Additional Results}\label{app:additional-results}

Here we extend our experiment from Section~\ref{sec:experiment} to additional language pairs and metrics of WMT 23. For the \emph{en-de} language pair $N^{H} = 460$ and $N^{M} = 557$ and for \emph{he-en} $N^{H} = 820$ and $H^{M} = 1910$. We show the results for \emph{XCOMET} for each language pair in Tables~\ref{tab:wmt23-ende-xcomet-scores},~\ref{tab:wmt23-heen-xcomet-scores}, and~\ref{tab:wmt23-zhen-xcomet-scores-repeat} (Note that Table~\ref{tab:wmt23-zhen-xcomet-scores-repeat} is the same as Table~\ref{tab:wmt23-zhen-xcomet-scores} in Section~\ref{sec:experiment}). We also include results for \emph{GEMBA-MQM}~\citep{gemba-mqm}, which is a reference free metric based on prompting LLMs. The results can be seen in Tables~\ref{tab:wmt23-ende-gembamqm-scores},~\ref{tab:wmt23-heen-gembamqm-scores}, and~\ref{tab:wmt23-zhen-gembamqm-scores}.

\begin{table*}[tbh]
    \centering
    \small
    \begin{tabular}{r | c c c c c c c}
 & \multicolumn{2}{c}{Human} & \multicolumn{2}{c}{Metric} & \multicolumn{2}{c}{Remapped} & Exp. Deviation \\
 & $\hat{\mu}_{k}^{H}$ & R & $\hat{\mu}_{k}^{M}$ & R & $\hat{\mu}_{k}^{G}$ & R & ED \\ \hline
GPT4-5shot & -3.724 & 1 & 0.882 & 2 & -4.768 & 1 & -1.044 \\
ONLINE-W & -3.950 & 2 & 0.883 & 1 & -4.821 & 2 & -0.871 \\
ONLINE-B & -4.711 & 3 & 0.871 & 3 & -5.272 & 3 & -0.560 \\
ONLINE-Y & -5.643 & 4 & 0.858 & 4 & -5.909 & 4 & -0.266 \\
ONLINE-A & -5.668 & 5 & 0.853 & 5 & -6.152 & 5 & -0.483 \\
ONLINE-G & -6.574 & 6 & 0.834 & 6 & -7.079 & 6 & -0.505 \\
ONLINE-M & -6.936 & 7 & 0.830 & 7 & -7.399 & 7 & -0.462 \\
Lan-BridgeMT & -8.670 & 8 & 0.801 & 9 & -8.670 & 9 & -0.000 \\
ZengHuiMT & -9.255 & 9 & 0.790 & 11 & -9.387 & 11 & -0.132 \\
NLLB-Greedy & -9.543 & 10 & 0.812 & 8 & -8.405 & 8 & 1.138 \\
NLLB-MBR-BLEU & -10.794 & 11 & 0.797 & 10 & -9.005 & 10 & 1.789 \\
AIRC & -14.228 & 12 & 0.724 & 12 & -13.658 & 12 & 0.570 \\
\end{tabular}

    \caption{System rankings and average rating of WMT 23 \emph{en-de} systems according to \emph{XCOMET}.}
    \label{tab:wmt23-ende-xcomet-scores}
\end{table*}

\begin{table*}[tbh]
    \centering
    \small
    \begin{tabular}{r | c c c c c c c}
 & \multicolumn{2}{c}{Human} & \multicolumn{2}{c}{Metric} & \multicolumn{2}{c}{Remapped} & Exp. Deviation \\
 & $\hat{\mu}_{k}^{H}$ & R & $\hat{\mu}_{k}^{M}$ & R & $\hat{\mu}_{k}^{G}$ & R & ED \\ \hline
GPT4-5shot & -1.333 & 1 & 0.913 & 2 & -1.690 & 2 & -0.358 \\
ONLINE-A & -1.381 & 2 & 0.908 & 3 & -1.817 & 3 & -0.436 \\
ONLINE-B & -1.546 & 3 & 0.916 & 1 & -1.635 & 1 & -0.089 \\
GTCOM-Peter & -1.886 & 4 & 0.904 & 4 & -1.916 & 4 & -0.030 \\
UvA-LTL & -1.919 & 5 & 0.893 & 6 & -2.193 & 6 & -0.274 \\
ONLINE-G & -2.055 & 6 & 0.895 & 5 & -2.137 & 5 & -0.082 \\
ONLINE-Y & -2.349 & 7 & 0.881 & 8 & -2.511 & 8 & -0.162 \\
ZengHuiMT & -2.382 & 8 & 0.889 & 7 & -2.294 & 7 & 0.088 \\
Samsung-Res.-Ph. & -3.234 & 9 & 0.874 & 9 & -2.666 & 9 & 0.568 \\
NLLB-MBR-BLEU & -3.678 & 10 & 0.869 & 11 & -2.805 & 11 & 0.872 \\
NLLB-Greedy & -3.790 & 11 & 0.872 & 10 & -2.714 & 10 & 1.076 \\
Lan-BridgeMT & -3.793 & 12 & 0.867 & 12 & -2.823 & 12 & 0.971 \\
\end{tabular}

    \caption{System rankings and average rating of WMT 23 \emph{he-en} systems according to \emph{XCOMET}.}
    \label{tab:wmt23-heen-xcomet-scores}
\end{table*}

\begin{table*}[tbh]
    \centering
    \small
    \begin{tabular}{r | c c c c c c c}
 & \multicolumn{2}{c}{Human} & \multicolumn{2}{c}{Metric} & \multicolumn{2}{c}{Remapped} & Exp. Deviation\\
 & $\hat{\mu}_{k}^{H}$ & R & $\hat{\mu}_{k}^{M}$ & R & $\hat{\mu}_{k}^{G}$ & R & ED \\ \hline
Lan-BridgeMT & -2.100 & 1 & 0.889 & 2 & -2.920 & 2 & \textit{-0.820} \\
GPT4-5shot & -2.305 & 2 & 0.893 & 1 & -2.800 & 1 & -0.494 \\
Yishu & -3.231 & 3 & 0.880 & 4 & -3.179 & 4 & 0.052 \\
ONLINE-B & -3.385 & 4 & 0.879 & 5 & -3.188 & 5 & 0.197 \\
HW-TSC & -3.398 & 5 & 0.883 & 3 & -3.080 & 3 & 0.318 \\
ONLINE-A & -3.785 & 6 & 0.856 & 8 & -3.812 & 8 & -0.027 \\
ONLINE-Y & -3.792 & 7 & 0.868 & 6 & -3.479 & 6 & 0.313 \\
ONLINE-G & -3.857 & 8 & 0.864 & 7 & -3.607 & 7 & 0.250 \\
ONLINE-W & -4.062 & 9 & 0.848 & 9 & -4.165 & 10 & -0.103 \\
ZengHuiMT & -4.232 & 10 & 0.846 & 10 & -4.140 & 9 & 0.092 \\
IOL-Research & -4.586 & 11 & 0.843 & 11 & -4.251 & 11 & 0.335 \\
ONLINE-M & -5.433 & 12 & 0.820 & 15 & -4.907 & 15 & 0.526 \\
ANVITA & -6.078 & 13 & 0.830 & 13 & -4.602 & 13 & 1.475 \\
NLLB-MBR-BLEU & -6.360 & 14 & 0.825 & 14 & -4.726 & 14 & 1.634 \\
NLLB-Greedy & -6.574 & 15 & 0.831 & 12 & -4.578 & 12 & \textbf{1.996} \\
\end{tabular}

    \caption{System rankings and average rating of WMT 23 \emph{zh-en} systems according to \emph{XCOMET}.}
    \label{tab:wmt23-zhen-xcomet-scores-repeat}
\end{table*}

\begin{table*}[tbh]
    \centering
    \small
    \begin{tabular}{r | c c c c c c c}
 & \multicolumn{2}{c}{Human} & \multicolumn{2}{c}{Metric} & \multicolumn{2}{c}{Remapped} & Exp. Deviation \\
 & $\hat{\mu}_{k}^{H}$ & R & $\hat{\mu}_{k}^{M}$ & R & $\hat{\mu}_{k}^{G}$ & R & ED \\ \hline
GPT4-5shot & -3.724 & 1 & -2.447 & 1 & -4.123 & 1 & -0.399 \\
ONLINE-W & -3.950 & 2 & -3.429 & 2 & -4.822 & 2 & -0.872 \\
ONLINE-B & -4.711 & 3 & -4.048 & 3 & -5.383 & 3 & -0.672 \\
ONLINE-Y & -5.643 & 4 & -4.424 & 4 & -5.832 & 5 & -0.189 \\
ONLINE-A & -5.668 & 5 & -4.567 & 5 & -5.826 & 4 & -0.158 \\
ONLINE-G & -6.574 & 6 & -6.018 & 6 & -7.047 & 6 & -0.473 \\
ONLINE-M & -6.936 & 7 & -6.217 & 7 & -7.113 & 7 & -0.177 \\
Lan-BridgeMT & -8.670 & 8 & -8.197 & 8 & -8.891 & 9 & -0.221 \\
ZengHuiMT & -9.255 & 9 & -8.357 & 9 & -8.867 & 8 & 0.388 \\
NLLB-Greedy & -9.543 & 10 & -10.043 & 10 & -9.683 & 10 & -0.140 \\
NLLB-MBR-BLEU & -10.794 & 11 & -10.724 & 11 & -10.352 & 11 & 0.442 \\
AIRC & -14.228 & 12 & -13.941 & 12 & -12.526 & 12 & 1.702 \\
\end{tabular}

    \caption{System rankings and average rating of WMT 23 \emph{en-de} systems according to \emph{GEMBA-MQM}.}
    \label{tab:wmt23-ende-gembamqm-scores}
\end{table*}

\begin{table*}[tbh]
    \centering
    \small
    \begin{tabular}{r | c c c c c c c}
 & \multicolumn{2}{c}{Human} & \multicolumn{2}{c}{Metric} & \multicolumn{2}{c}{Remapped} & Exp. Deviation \\
 & $\hat{\mu}_{k}^{H}$ & R & $\hat{\mu}_{k}^{M}$ & R & $\hat{\mu}_{k}^{G}$ & R & ED \\ \hline
GPT4-5shot & -1.333 & 1 & -1.923 & 1 & -1.377 & 1 & -0.045 \\
ONLINE-A & -1.381 & 2 & -3.850 & 2 & -1.882 & 2 & -0.501 \\
ONLINE-B & -1.546 & 3 & -4.108 & 3 & -1.969 & 3 & -0.423 \\
GTCOM-Peter & -1.886 & 4 & -4.859 & 4 & -2.144 & 4 & -0.258 \\
UvA-LTL & -1.919 & 5 & -5.628 & 6 & -2.312 & 6 & -0.393 \\
ONLINE-G & -2.055 & 6 & -5.240 & 5 & -2.281 & 5 & -0.225 \\
ONLINE-Y & -2.349 & 7 & -6.885 & 8 & -2.677 & 8 & -0.328 \\
ZengHuiMT & -2.382 & 8 & -6.032 & 7 & -2.484 & 7 & -0.102 \\
Samsung-Res.-Ph. & -3.234 & 9 & -8.545 & 12 & -2.954 & 12 & 0.280 \\
NLLB-MBR-BLEU & -3.678 & 10 & -8.075 & 9 & -2.817 & 10 & 0.861 \\
NLLB-Greedy & -3.790 & 11 & -8.261 & 10 & -2.813 & 9 & 0.977 \\
Lan-BridgeMT & -3.793 & 12 & -8.469 & 11 & -2.840 & 11 & 0.953 \\
\end{tabular}

    \caption{System rankings and average rating of WMT 23 \emph{he-en} systems according to \emph{GEMBA-MQM}.}
    \label{tab:wmt23-heen-gembamqm-scores}
\end{table*}

\begin{table*}[tbh]
    \centering
    \small
    \begin{tabular}{r | c c c c c c c}
 & \multicolumn{2}{c}{Human} & \multicolumn{2}{c}{Metric} & \multicolumn{2}{c}{Remapped} & Exp. Deviation \\
 & $\hat{\mu}_{k}^{H}$ & R & $\hat{\mu}_{k}^{M}$ & R & $\hat{\mu}_{k}^{G}$ & R & ED \\ \hline
Lan-BridgeMT & -2.100 & 1 & -1.949 & 2 & -2.419 & 2 & -0.319 \\
GPT4-5shot & -2.305 & 2 & -1.601 & 1 & -2.199 & 1 & 0.106 \\
Yishu & -3.231 & 3 & -4.790 & 5 & -3.492 & 5 & -0.261 \\
ONLINE-B & -3.385 & 4 & -4.717 & 4 & -3.489 & 4 & -0.104 \\
HW-TSC & -3.398 & 5 & -4.367 & 3 & -3.336 & 3 & 0.062 \\
ONLINE-A & -3.785 & 6 & -5.568 & 8 & -3.838 & 9 & -0.053 \\
ONLINE-Y & -3.792 & 7 & -5.453 & 7 & -3.611 & 6 & 0.181 \\
ONLINE-G & -3.857 & 8 & -5.275 & 6 & -3.724 & 7 & 0.134 \\
ONLINE-W & -4.062 & 9 & -5.760 & 9 & -3.772 & 8 & 0.290 \\
ZengHuiMT & -4.232 & 10 & -6.337 & 10 & -4.089 & 11 & 0.143 \\
IOL-Research & -4.586 & 11 & -6.511 & 11 & -4.067 & 10 & 0.519 \\
ONLINE-M & -5.433 & 12 & -9.115 & 12 & -4.899 & 13 & 0.534 \\
ANVITA & -6.078 & 13 & -9.440 & 13 & -4.844 & 12 & 1.234 \\
NLLB-MBR-BLEU & -6.360 & 14 & -11.339 & 15 & -5.379 & 15 & 0.981 \\
NLLB-Greedy & -6.574 & 15 & -11.282 & 14 & -5.312 & 14 & 1.262 \\
\end{tabular}

    \caption{System rankings and average rating of WMT 23 \emph{zh-en} systems according to \emph{GEMBA-MQM}.}
    \label{tab:wmt23-zhen-gembamqm-scores}
\end{table*}

\section{Evaluating the System Dependence of WMT23 Metrics}\label{sec:sysdep}

In Section~\ref{sec:theory}, we introduced the \textit{SysDep} score. It measures the worst case in the difference of expected deviations (ED), which measures the difference between the average human rating we expect to see based on metric ratings and assuming a single global $f_G$ and the true average human rating for a system $\pi_k$. To measure the system dependence of a metric across a set of systems $\pi_1, \dots, \pi_K$, we compute the range of the individual $ED$: $SysDep = max_{\pi_k} ED(k) - min_{\pi_k} ED(k)$. We noted in Section~\ref{sec:experiment} that $ED(k)$ alone is not enough to know whether system $\pi_k$ will be ranked incorrectly, it depends on the true margin to the other systems, and their dependencies. By measuring the range, we consider the worst case when comparing two systems. We show the dependency ranges for all WMT23 metrics on all language pairs in Table~\ref{tab:sysdeps}. We notice that the values for \emph{en-de} are large than the others. This is due to a larger range of human rating averages for this language pair (see also Tables~\ref{tab:wmt23-ende-xcomet-scores}--\ref{tab:wmt23-zhen-gembamqm-scores} in Appendix~\ref{app:additional-results}). We therefore also do not aggregate across language pairs.

\begin{table}[tbh]
    \small
    \begin{tabular}{ l | c c c}
 & en-de & he-en & zh-en \\ \hline
BERTscore & 7.18 & 1.73 & 3.87 \\
BLEU & 9.02 & 2.06 & 4.23 \\
BLEURT-20 & 3.68 & 1.66 & 3.35 \\
Calibri-COMET22-QE & 3.68 & 2.06 & 3.30 \\
Calibri-COMET22 & 4.24 & 1.64 & 3.40 \\
chrF & 8.11 & 1.92 & 4.29 \\
COMET & 4.29 & 1.64 & 3.35 \\
CometKiwi & 3.84 & 1.95 & 3.02 \\
CometKiwi-XL & 3.77 & 1.98 & 3.01 \\
CometKiwi-XXL & 3.68 & 1.82 & 2.92 \\
cometoid22-wmt21 & 5.44 & 2.11 & 3.30 \\
cometoid22-wmt22 & 5.17 & 2.09 & 3.21 \\
cometoid22-wmt23 & 4.66 & 1.81 & 3.20 \\
docWMT22CometDA & 3.87 & 1.65 & 3.37 \\
docWMT22CometKiwiDA & 4.53 & 1.83 & 2.76 \\
eBLEU & 9.49 & 2.08 & 4.29 \\
embed-llama & 7.07 & 2.13 & 4.21 \\
f200spBLEU & 8.42 & 2.01 & 4.23 \\
GEMBA-MQM & 2.57 & 1.48 & \bf{1.58} \\
instructscore & 3.59 & 1.53 & 3.68 \\
KG-BERTScore & 4.24 & 1.88 & 3.04 \\
MaTESe & 5.98 & 1.49 & 3.16 \\
mbr-metricx-qe & 3.69 & 1.58 & 2.39 \\
MEE4 & 8.48 & 1.88 & 4.21 \\
MetricX-23-b & 2.26 & 1.29 & 2.81 \\
MetricX-23-c & 3.56 & 1.69 & 2.36 \\
MetricX-23-QE-b & \bf{2.11} & 1.55 & 2.62 \\
MetricX-23-QE-c & 2.82 & \bf{1.21} & 1.65 \\
MetricX-23-QE & 2.93 & 1.77 & 3.12 \\
MetricX-23 & 2.57 & 1.33 & 3.04 \\
mre-score-labse-regular & 9.70 & 1.65 & 3.94 \\
MS-COMET-QE-22 & 5.87 & 2.22 & 3.37 \\
prismRef & 8.71 & 1.79 & 3.97 \\
prismSrc & \it{11.24} & 2.48 & \it{4.61} \\
Random-sysname & 9.97 & \it{2.52} & 4.53 \\
sescoreX & 3.59 & 1.52 & 3.47 \\
tokengram-F & 8.17 & 1.93 & 4.29 \\
XCOMET-Ensemble & 2.83 & 1.51 & 2.82 \\
XCOMET-QE-Ensemble & 2.95 & 1.78 & 2.95 \\
XCOMET-XL & 3.39 & 1.59 & 3.20 \\
XCOMET-XXL & 2.71 & 1.48 & 2.99 \\
XLsim & 7.83 & 2.01 & 4.20 \\
YiSi-1 & 5.95 & 1.60 & 3.65 \\
\end{tabular}

    \caption{$SysDep$ for each metric and language pair. We show the \textbf{minimum} and \textit{maximum} for each language pair.}
    \label{tab:sysdeps}
\end{table}

Variants of \emph{MetricX-23}~\citep{metricx} perform best on \emph{en-de} and \emph{he-en}, while \emph{GEMBA-MQM} has the lowest range for \emph{zh-en}. The reference-free \emph{prismSrc}~\citep{prismA, prismB} metric performs worst on \emph{en-de} and \emph{zh-en}. The baseline \emph{Random-sysname}~\citep{wmt_metrics_23} performs worst for \emph{he-en}. This baseline is an interesting case, as it is the prototypical example of a metric where every $f_k$ is different. It assigns a fixed score to each system based on its name and adds Gaussian noise to this value to assign segment level scores. Therefore each $f_k$ will be a different constant function.

\section{Intra-System Variability}\label{sec:intra-system}

In order to confirm that the observed \textit{SysDep} scores are indeed due to a metric systematically treating systems differently and not due to variance in ratings, we will measure the maximum intra-system scores. For this, we use ratings from a single system and split them into $2$ equal sized parts $10$ times with different random seeds. This simulates a setting with $20$ systems with half the sample size of the original setting. We then compute the \textit{SysDep} score. 

In Table~\ref{tab:intra-system}, we show the maximum intra-system \textit{SysDep} score computed this way over all systems for a given metric and language pair. We observe that for \emph{he-en} and \emph{zh-en} all scores are lower than the minimum between system \textit{SysDep} reported in Appendix~\ref{sec:sysdep}. This confirms that in those cases metrics treat different systems differently.
For the \emph{en-de} language pair, we observe that while in many cases the intra-system score is lower than the \textit{SysDep} between systems for the same metric and language pair, this is not always the case. This could be due to the metrics treating systems more equally for this language pair, or the relatively small sample sizes for \emph{en-de} compared to the other language pairs.

\begin{table}[tbh]
    \small
    \begin{tabular}{ l | c c c}
 & en-de & he-en & zh-en \\ \hline
BERTscore & 2.08 & 0.88 & 1.00 \\
BLEU & 2.20 & 0.95 & 0.99 \\
BLEURT-20 & 2.68 & 0.89 & 0.87 \\
Calibri-COMET22-QE & 2.97 & 0.82 & 0.85 \\
Calibri-COMET22 & 2.50 & 0.79 & 0.89 \\
chrF & 2.07 & 0.96 & 0.93 \\
COMET & 2.25 & 0.78 & 0.87 \\
CometKiwi & 3.01 & 0.83 & 0.81 \\
CometKiwi-XL & 2.90 & 0.87 & 0.80 \\
CometKiwi-XXL & 2.65 & 0.83 & 0.85 \\
cometoid22-wmt21 & 2.81 & 0.76 & 0.89 \\
cometoid22-wmt22 & 2.74 & 0.73 & 0.81 \\
cometoid22-wmt23 & 2.58 & 0.79 & 0.86 \\
docWMT22CometDA & 2.32 & 0.83 & 0.95 \\
docWMT22CometKiwiDA & 2.61 & 0.88 & 0.95 \\
eBLEU & 2.49 & 0.91 & 0.98 \\
embed-llama & 2.18 & 0.92 & 1.14 \\
f200spBLEU & 2.10 & 0.94 & 0.95 \\
GEMBA-MQM & 2.89 & 0.93 & 0.88 \\
instructscore & 2.20 & 0.83 & 0.82 \\
KG-BERTScore & 2.90 & 0.85 & 0.82 \\
MaTESe & 2.71 & 0.77 & 0.80 \\
mbr-metricx-qe & 2.52 & 0.84 & 0.79 \\
MEE4 & 2.15 & 0.91 & 0.94 \\
MetricX-23-b & 2.67 & 0.87 & 0.68 \\
MetricX-23-c & 2.89 & 0.90 & 0.77 \\
MetricX-23-QE-b & 2.67 & 0.80 & 0.71 \\
MetricX-23-QE-c & 2.25 & 0.80 & 0.81 \\
MetricX-23-QE & 2.46 & 0.80 & 0.74 \\
MetricX-23 & 2.34 & 0.90 & 0.63 \\
mre-score-labse-regular & 2.27 & 0.91 & 0.93 \\
MS-COMET-QE-22 & 2.17 & 0.93 & 0.90 \\
prismRef & 2.24 & 0.83 & 0.99 \\
prismSrc & 2.09 & 0.88 & 0.91 \\
Random-sysname & 2.36 & 0.96 & 0.95 \\
sescoreX & 2.20 & 0.86 & 0.86 \\
tokengram-F & 2.10 & 0.96 & 0.94 \\
XCOMET-Ensemble & 2.44 & 0.71 & 0.68 \\
XCOMET-QE-Ensemble & 2.30 & 0.72 & 0.72 \\
XCOMET-XL & 2.44 & 0.74 & 0.67 \\
XCOMET-XXL & 2.35 & 0.74 & 0.69 \\
XLsim & 2.71 & 0.90 & 0.94 \\
YiSi-1 & 2.44 & 0.83 & 0.90 \\
\end{tabular}

    \caption{Maximum intra-system \textit{SysDep} score for all metrics and language pairs.}
    \label{tab:intra-system}
\end{table}

\section{Estimating Conditional Expectations}\label{sec:code}

In Section~\ref{sec:experiment}, we gave a brief overview of how to compute estimates for the functions $f_k$ and $f_G$. In Listings~\ref{lst:iso_part1} and~\ref{lst:iso_part2}, we show our \textit{python} implementation. To estimate the system-level $\hat{f}_k$, we call the \textit{.fit} method with human and metric ratings for system $\pi_k$. To evaluate the function $\hat{f}_k$, we use the \textit{.conditional\_expectation} method. To estimate the global function $\hat{f}_G$, we use the \textit{.fit} method with paired human and metric ratings for all systems. We compute the remapped rating $\hat{\mu}^{G}_{k}$ by first fitting $\hat{f}_G$ and then using the \textit{.remapped\_expectation} method on the metric ratings for system $\pi_k$. We rely on the \textit{Isotonic Regression} implementation from \textit{scikit-learn}~\citep{scikit-learn}~\footnote{\url{https://scikit-learn.org/stable/modules/generated/sklearn.isotonic.IsotonicRegression.html}} and numerical utility functions from \textit{numpy}~\citep{numpy}.

\begin{figure*}[tb]
    \centering
    \begin{minipage}{\textwidth}
\begin{lstlisting}[caption={Part 1 of the \textit{python} code to estimate $\hat{f}_k$, $\hat{f}_G$, and $\hat{\mu}^{G}_{k}$.}, label={lst:iso_part1}]

from typing import Self, Tuple

import numpy as np
from numpy.random import Generator, default_rng

from sklearn.isotonic import IsotonicRegression


class BootstrapIsotonic:

    def __init__(
        self,
        n_bootstrap: int = 200,
        rng: int | Generator = 0xdeadbeef,
    ):
        self.n_bootstrap = n_bootstrap
        self.rng = default_rng(rng)
        self.models = []

    def fit(
        self,
        human_ratings: np.ndarray[float],  # 1d array of human ratings
        metric_ratings: np.ndarray[float], # 1d array of matching metric ratings
    ) -> Self:
        # fit a model of f_k or f_G, i.e. the conditional expectation of human ratings given metric ratings
        # to get model a system-level function f_k, use only human_ratings for that given system k
        # to model the global function f_G, use data from all systems

        assert len(human_ratings) == len(metric_ratings)
        n_samples = len(human_ratings)

        for _ in range(self.n_bootstrap):
            bootstrap_indices = self.rng.choice(
                np.arange(n_samples),
                n_samples,
                replace=True
            )
            isotonic_model = IsotonicRegression(
                y_min=None,  # MQM scores range from large negative to 0
                y_max=0.,
                increasing=True,  # metric has positive correlation
                out_of_bounds='nan',  # don't extrapolate
            )
            isotonic_model.fit(
                X=metric_ratings[bootstrap_indices],
                y=human_ratings[bootstrap_indices],
            )
            self.models.append(isotonic_model)

        return self

    def _predict_bootstrap(
        self,
        m: np.ndarray[float]  # 1d array of metric ratings
    ) -> np.ndarray[float]:
        # helper function getting predictions from each model, returns 2d array of size [n_bootstrap, len(m)]
        result = np.zeros((self.n_bootstrap, len(m)), dtype=float)
        for bix, model in enumerate(self.models):
            result[bix, :] = model.predict(m)
        return result

\end{lstlisting}
    \end{minipage}
\end{figure*}

\begin{figure*}[tb]
    \centering
    \begin{minipage}{\textwidth}
\begin{lstlisting}[caption={Part 2 of the \textit{python} code to estimate $\hat{f}_k$, $\hat{f}_G$, and $\hat{\mu}^{G}_{k}$.}, label={lst:iso_part2}]

    def conditional_expectation(
        self,
        metric_ratings: np.ndarray[float],  # 1d array of metric ratings
    ) -> np.ndarray[float]:
        # this computes the function f_k or f_G (depending on what data we fitted on)
        bootstrap_predictions = self._predict_bootstrap(metric_ratings)
        return np.nanmean(bootstrap_predictions, axis=0)

    def confidence(
        self,
        metric_ratings: np.ndarray[float],  # 1d array of metric ratings
    ) -> Tuple[np.ndarray[float], np.ndarray[float]]:
        # this computes the confidence bounds around f_k or f_G in Figure 1
        bootstrap_predictions = self._predict_bootstrap(metric_ratings)
        lower = np.nanpercentile(bootstrap_predictions, 2.5, axis=0)
        upper = np.nanpercentile(bootstrap_predictions, 97.5, axis=0)
        return lower, upper

    def remapped_expectation(
        self,
        metric_ratings: np.ndarray[float], # 1d array of metric ratings
    ) -> float:
        # used to compute remapped system scores in Table 1.
        expected_human_ratings = self.conditional_expectation(metric_ratings)
        return np.nanmean(expected_human_ratings)
\end{lstlisting}
    \end{minipage}
\end{figure*}

\end{document}